\documentclass{article}
\usepackage{amssymb}
\usepackage{color}
\usepackage{tabularx}
\usepackage{multirow} 
\usepackage[dvipsnames]{xcolor}

\usepackage{multicol}
\usepackage{graphicx}
\usepackage{stfloats}
\usepackage[bookmarks=true]{hyperref}
\usepackage{times}
\usepackage{graphicx}
\usepackage{array}
\usepackage{booktabs}
\usepackage{amsmath,amsfonts}
\usepackage{siunitx}
\usepackage{diagbox}

\usepackage{multicol}
\usepackage[bookmarks=true]{hyperref}

\usepackage[final]{corl_2025}

\title{Towards Generalizable Safety in Crowd Navigation via Conformal Uncertainty Handling}

\author{
  Jianpeng Yao, Xiaopan Zhang\thanks{Equal contribution; $^{\dagger}$Corresponding author}\, , Yu Xia$^{*}$, Zejin Wang,\\ 
  \textbf{Amit K. Roy-Chowdhury, and Jiachen Li$^{\dagger}$}\\ \\
  University of California, Riverside \\
  \texttt{\{jyao073, jiachen.li\}@ucr.edu}
}

\begin{document}
\maketitle

\begin{abstract}
Mobile robots navigating in crowds trained using reinforcement learning are known to suffer performance degradation when faced with out-of-distribution scenarios. We propose that by properly accounting for the uncertainties of pedestrians, a robot can learn safe navigation policies that are robust to distribution shifts.  
Our method augments agent observations with prediction uncertainty estimates generated by adaptive conformal inference, and it uses these estimates to guide the agent’s behavior through constrained reinforcement learning. The system helps regulate the agent’s actions and enables it to adapt to distribution shifts. In the in-distribution setting, our approach achieves a 96.93\% success rate, which is over 8.80\% higher than the previous state-of-the-art baselines with over 3.72 times fewer collisions and 2.43 times fewer intrusions into ground-truth human future trajectories. In three out-of-distribution scenarios, our method shows much stronger robustness when facing distribution shifts in velocity variations, policy changes, and transitions from individual to group dynamics. We deploy our method on a real robot, and experiments show that the robot makes safe and robust decisions when interacting with both sparse and dense crowds. Our code and videos are available on \url{https://gen-safe-nav.github.io/}.
\end{abstract}


\section{Introduction}
Safely navigating among crowds is fundamental to a future where robots work closely with humans \cite{francis2023principles, mavrogiannis2023core, zhou2022grouptron, xu2024matrix}. Various solutions have been studied, including rule-based methods \cite{van2008reciprocal, helbing1995social}, optimization-based planners \cite{han2025dr, samavi2024sicnav, jansma2023interaction, li2023game}, reinforcement learning (RL) approaches \cite{chen2017decentralized, ma2021reinforcement, liu2021decentralized, kim2025human}, and hybrid systems that track a reference path from an optimizer while a learned policy makes necessary adaptations \cite{xie2023drl, xie2024scope}. 
For RL approaches, a single millisecond-level forward pass can generate decisions, making them faster and more scalable than optimization- or rule-based planners. Yet while they excel in in-distribution settings, their performance drops sharply in out-of-distribution (OOD) environments \cite{lee2023robust,arief2024importance}. This indicates a tendency towards overfitting and difficulty in generalizing to diverse crowd dynamics \cite{ma2021multi}.

Moreover, previous studies have shown that incorporating predictions into observations, thus forming prediction-augmented observations \cite{li2021spatio, liu2023intention, ma2021continual, xu2023safecrowdnav, li2024interactive, wang2025deployable, wang2025uniocc}, can explicitly represent human intentions \cite{girase2021loki, li2023pedestrian} and aid in robot decision making. However, this practice may exacerbate overfitting issues, as human dynamics in real-world scenarios are inherently complex and difficult to fully capture in simulation environments or datasets, and trajectory predictions fitted for one certain dynamics face generalization challenges to other dynamics. Inaccurate predictions can severely mislead robot decisions, especially when robots rely heavily on them to determine actions. Due to compounded overfitting from both predictions and the learned policy, robots specialized for these environments often fail to generalize to new scenarios with different crowd dynamics. In other words, existing methods lack a systematic treatment of prediction errors to improve policy generalizability.

In this paper, we propose that by properly quantifying uncertainties in human trajectory predictions and incorporating the results into decision making algorithms, it is possible to alleviate overfitting in RL-based crowd navigation. Uncertainty serves as an indicator of the reliability of prediction hypotheses, reflecting both prediction errors and the sensitivity of prediction models to distribution shifts \cite{li2024adaptive}. Therefore, encouraging the agent to account for uncertainty enables it to generate actions that are more robust to distribution shifts and resilient to incorrect assumptions about human dynamics. Specifically, we introduce a learning-based framework that explicitly reasons about prediction uncertainty. First, we apply adaptive conformal inference (ACI) \cite{gibbs2021adaptive,gibbs2024conformal} to quantify the uncertainty of each predicted human trajectory with a prediction set that contains the true future position with a user‑defined coverage probability. Unlike other conformal approaches \cite{angelopoulos2023conformal}, ACI updates its calibration online, so it can swiftly adapt when the underlying crowd dynamics shift. Second, we employ constrained reinforcement learning (CRL) to introduce effective controllability into the decision making system, using uncertainty estimates to guide both the learning process and the agent’s behavior. Our system achieves the state-of-the-art (SOTA) performance in safety metrics and with much smaller performance drops in OOD settings. In the in-distribution setting, our system achieves an over 8.80\% higher success rate than previous SOTA RL baselines, with more than 3.72 times fewer collisions and over 2.43 times fewer intrusions into pedestrian trajectories in in-distribution settings. In three different OOD test scenarios that introduce velocity shifts, policy shifts, and pedestrian grouping behavior, our method maintains a high success rate and low collision rate, while competing approaches degrade significantly. Finally, we deploy the learned policy on a real Mecanum-wheel robot. With only minor clipping and smoothing, the policy transfers directly from simulation and achieves safe navigation in both sparse and dense crowds. 

\section{Related Work}
\textbf{Crowd Navigation.} Mobile robots are expected to interact with humans and complete various tasks, such as providing assistance \cite{francis2023principles}. Crowd navigation forms the foundation for performing most high-level tasks. Robots are required to navigate in crowds, where the challenge of modeling dynamic human behavior makes navigation challenging. It is crucial to capture the subtleties of human behavior, such as human intentions and interactions between agents \cite{sun2022interaction,li2020evolvegraph,zhou2022grouptron,dax2023disentangled,lange2023scene,wang2025cmp}, and properly use them for effective robotic decisions.
Deep RL offers a potentially viable solution to the challenging navigation task \cite{li2023game,luo2025bridging}.
Previous work on RL-based methods for social robots includes the capture of agent-agent interactions \cite{chen2019crowd,li2024multi} and the intentions of human agents \cite{liu2023intention,ma2022multi}, incorporating these as predictions into RL policy networks. Our work focuses on alleviating performance degradation in OOD scenarios and proposes a system that works towards this challenge by effectively considering and handling uncertainties in human behavior predictions.

\textbf{Planning Under Uncertainty.}
Trajectory planning under uncertainty has attracted increasing attention in the past decades. In optimization-based and search-based approaches, researchers have integrated uncertainty quantification from perception and prediction into various controllers \cite{yang2023safe, dixit2023adaptive, huang2025interaction, lindemann2023safe, wang2025generative}. These methods easily support adding constraints or safety shields, allowing explicit management of uncertainty.
By contrast, guiding decision making in RL agents with uncertainty measures is less straightforward, since policy networks often behave as black boxes. Prior work has augmented agent observations with uncertainty estimates \cite{huang2023conformal, golchoubian2024uncertainty} or post-processed policy outputs based on uncertainty to ensure safe behavior \cite{strawn2023conformal}. However, most approaches avoid integrating uncertainty guidance directly into the learning process and still suffer from OOD performance degradation similar to the methods without uncertainty measures.
In our work, we propose that uncertainty-aware planning can effectively mitigate OOD performance degradation for RL-based crowd navigation and introduce a CRL framework that enhances controllability and leverages uncertainty estimates to effectively guide decision making, thereby improving robustness to prediction errors and OOD scenarios.

\section{Method}
\subsection{Problem Formulation}
In our setting, we have $H$ humans in the environment, each indexed by $h$, within an episode of horizon $T$. 
At each time step $t$, the positions of humans are represented as $\mathbf{p}_{h}(t)$. We predict $K$ future steps (a larger $K$ means more extended future predictions) for each human's future trajectories. The prediction point for the $k$-th prediction of the $h$-th human is denoted as $\mathbf{p}_{h, k}(t)$, where $1 \leq h \leq H$ and $1 \leq k \leq K$. We formulate the task as a constrained Markov decision process, where the CRL agent is provided with observations of the state \(S_t\) at each timestep, which consists of two main parts. The first part includes physical information: the current positions of humans and the robot, and other quantities about the robot's dynamics. The second part comprises post-processed features generated by models, such as the predicted human trajectories. We denote physical state components of ego information as \(\mathbf{e}\), physical components of human information as \(\mathbf{h}\), and components generated by models as \(\mathbf{m}\). The complete state \(S_t\) is written as \(S_t = [\mathbf{e}, \mathbf{h}, \mathbf{m}]\).
The agent then generates action \(A_t = (v_x, v_y)\) to control the moving speed of the robot according to \(S_t\).
After taking an action, the environment provides a reward \(R_t\) and a cost \(C_t\). We aim to obtain an optimal policy \(\pi(A_t \mid S_t)\) that maximizes rewards while satisfying cost constraints.

\begin{figure*}[!tbp]
	\centering
	\includegraphics[width=1.0\textwidth]{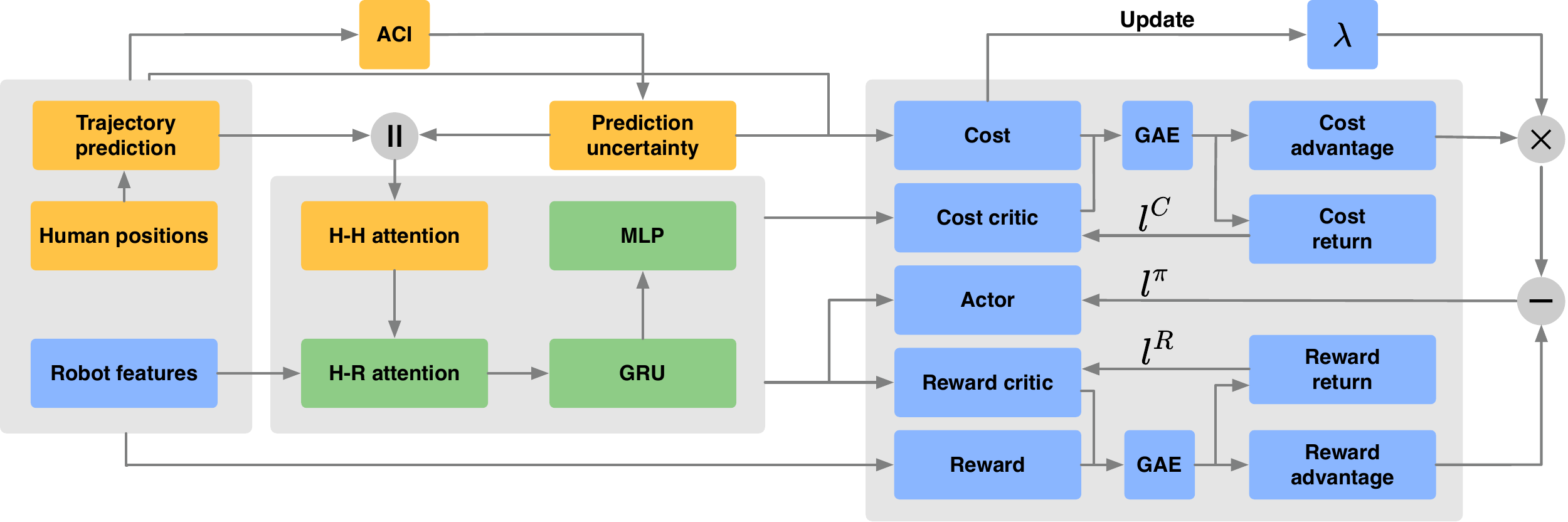}
	\caption{The overall pipeline of our method. We mark components related to humans in yellow, components related to physical information and decision making of the robot in blue, and fused features in green. We use ACI to quantify the prediction uncertainty of human trajectories and concatenate these metrics with predictions before inputting them into networks. The networks contain attention mechanisms for interactions between humans (H-H attention) and between humans and the ego robot (H-R attention). Prediction uncertainty combined with physical information is used for designing costs. For the CRL agent using PPO Lagrangian, the actor and reward critic share some layers while the cost critic uses a separate network. We adopt reward value loss $l^R$, action loss $l^{\pi}$, and cost value loss $l^C$ for updating the agent.}
	\label{fig:overall_diagram}
\end{figure*}

\subsection{Method Overview}
The overall pipeline of our method is illustrated in Fig.~\ref{fig:overall_diagram}. First, we employ two different trajectory predictors: the constant velocity (CV) predictor \cite{scholler2020constant} and the Gumbel social transformer (GST) predictor \cite{huang2021learning}. This is to demonstrate that our algorithm can adapt to both rule-based and learning-based predictors. Next, we use ACI to quantify prediction uncertainties and incorporate them into \(\mathbf{m}\); subsequently, we adopt a policy network with a combined attention mechanism as proposed in \cite{liu2023intention} to process these uncertainties along with other features. Finally, we employ CRL to guide the agents' behavior using the uncertainty quantification results. Instead of directly applying constraints to the collision rate, we impose constraints on the cumulative intrusions of the robot into other agents’ uncertainty areas. This approach provides behavior-level guidance and effectively addresses the issue of sparse constraint feedback, thereby improving upon previous methods that constrained result-oriented metrics such as the collision rate.

\subsection{Rule-Based and Learning-Based Trajectory Prediction}
Our system adapts to different prediction models, including both learning-based and rule-based trajectory predictors, and can mitigate the adverse effects of incorrect predictions on the subsequent decision making process. For the rule-based prediction model, we use the CV model to obtain simple and effective estimates of future human states by extrapolating the current states of human agents based on their velocities.
For learning-based prediction models, we choose the GST predictor \cite{huang2021learning} as the learning-based prediction model, which is designed to address the challenges of partially detected pedestrians and redundant interaction modeling. GST is adapted for the efficient encoding of pedestrian features and demonstrates adaptability and robustness in high-density crowds.

\subsection{Adaptive Conformal Inference for Quantifying Prediction Uncertainty}
After obtaining the $K$-step future predictions, we quantify the prediction uncertainty using dynamically-tuned adaptive conformal inference (DtACI) \cite{gibbs2024conformal}, an ACI algorithm that adapts effectively to distribution shifts and thus serves as a good fit for online uncertainty estimation in social navigation. For each prediction step of each pedestrian, we run $M$ prediction error estimators simultaneously. At time step $t$, we calculate the actual prediction error $\delta_{h, k}$ between the current position and the predicted position made at time step $t-k$ for the $k$-th prediction step of the $h$-th human as $\delta_{h, k}(t) = \| \mathbf{p}_{h}(t) - \mathbf{p}_{h, k}(t-k) \|_2$, where $\mathbf{p}_{h}(t)$ is the position of the $h$-th human at time $t$, $\mathbf{p}_{h, k}(t-k)$ is the $k$-step predicted position of the $h$-th human made at time $t-k$, and $\delta_{h, k}(t)$ is calculated as the L2 norm of the difference of the two values. We then update the estimated error generated by the $m$-th estimator for the $h$-th human as
\begin{equation} 
    \hat{\delta}^{(m)}_{h, k}(t) = \hat{\delta}^{(m)}_{h, k}(t-1) 
    \;-\; \gamma^{(m)} \bigl(\alpha - \operatorname{err}_{h, k}^{(m)}(t)\bigr),
\end{equation}
where $\hat{\delta}^{(m)}_{h, k}$ represents the estimated prediction error of the $m$-th estimator corresponding to a $k$-step ahead prediction for the $h$-th human, $\gamma^{(m)}$ is the learning rate of the $m$-th prediction error estimator for all humans and predictions, $\alpha$ is the coverage parameter, and 
\begin{equation}
    \operatorname{err}^{(m)}_{h, k}(t) := 
    \begin{cases} 
        1, & \text{if } \hat{\delta}^{(m)}_{h, k}(t-1) < \delta_{h, k}(t), \\ 
        0, & \text{if } \hat{\delta}^{(m)}_{h, k}(t-1) \geq \delta_{h, k}(t).
    \end{cases}
\end{equation}
According to quantile regression \cite{gibbs2024conformal}, $\hat{\delta}^{(m)}_{h, k}$ converges to the $(1-\alpha)$ quantile of actual errors.
Since we run $M$ prediction error estimators with different learning rates simultaneously, for each error estimator, after taking in the actual prediction error $\delta_{h, k}$ and updating the estimated prediction error for the next step, we evaluate the errors of each estimator and update the probability distribution for choosing the next output prediction error estimator by
\begin{equation}
    w_{h, k}^{(m)} \leftarrow (1 - \sigma)\,\frac{w_{h, k}^{(m)} 
    \exp\bigl(-\eta\, \ell(\delta_{h, k}, \delta_{h, k}^{(m)})\bigr)}
    {\sum_{j=1}^{M} w_{h, k}^{(j)} 
    \exp\bigl(-\eta\,\ell(\delta_{h, k}, \delta_{h, k}^{(j)})\bigr)} 
    \;+\; \frac{\sigma}{M}, \quad
    p_{h, k}^{(m)} \leftarrow \frac{w_{h, k}^{(m)}}
    {\sum_{j=1}^{M} w_{h, k}^{(j)}},
\end{equation}
where we ignore the explicit time-step notation $t$ for simplicity, and use the arrow to indicate the replacement of values. The weight $w_{h, k}^{(m)}$ corresponds to the probability $p_{h, k}^{(m)}$ for the $m$-th prediction error estimator, $\sigma$ and $\eta$ are hyperparameters of DtACI for adjusting the speed at which weights change. $\ell(\delta_{h, k}, \delta_{h, k}^{(m)})$ is the pinball loss function used to measure the estimation error:
\begin{equation}
    \ell(\delta_{h, k}, \delta_{h, k}^{(m)}) =
    \begin{cases}
        \alpha \,\bigl(\delta_{h, k} - \delta_{h, k}^{(m)}\bigr), 
            & \text{if } \delta_{h, k} \ge \delta_{h, k}^{(m)},\\
        (\alpha - 1)\,\bigl(\delta_{h, k} - \delta_{h, k}^{(m)}\bigr), 
            & \text{if } \delta_{h, k} < \delta_{h, k}^{(m)}.
    \end{cases}
\end{equation}
Each time we estimate the prediction uncertainty, we treat
$\hat{\delta}_{h, k}$ as a discrete random variable taking values in the set
$\{\hat{\delta}_{h, k}^{(1)}, \ldots, \hat{\delta}_{h, k}^{(M)}\}$, with probability mass function taken values from $\{p_{h, k}^{(m)}\}$.

\subsection{Policy Network Structure}
Once we have obtained the trajectory prediction results and the corresponding prediction uncertainty, we concatenate the uncertainty quantification with the predicted trajectory before feeding it into the attention layers. This allows the RL agents to account for the prediction uncertainty in their decision making process, as shown in Fig.~\ref{fig:overall_diagram}. The first block is the human-human attention (H-H attention in Fig.~\ref{fig:overall_diagram}), which models each human as a separate node to capture interactions among humans. Next, we fuse the robot features (including velocity, heading, positions, and goal) into the attention blocks through human-robot attention (H-R attention) to obtain fused feature embeddings capturing the interactions between humans and the ego robot. We then process these fused embeddings with the robot features and concatenate them to form the GRU input, thereby capturing temporal information. Lastly, the final fused features are passed to the actor and critic networks for further processing. For more details about the policy network structure, please refer to~\cite{liu2023intention}.

\subsection{Decision Making with Effective Uncertainty Handling}
To enhance controllability and enable effective guidance under uncertainty, we adopt a physically meaningful cost and adjust its distribution based on a predefined cost limit using CRL, which is not achievable with traditional RL and reward shaping, as conventional rewards provide only an aggregate signal without any mechanism to ensure that safety-related components converge within a specified range. First, we design the safety critical area of pedestrians as a combination of a circular area around the human's position and an uncertainty area around $K^\prime$ ($K^\prime \leq K$) steps of predictions. Since we have $H$ human agents in the environment, the two parts of the area are defined as
\begin{equation}
D_i(\mathbf{p}_{\text{ego}}) = \left\{ \mathbf{p}_{\text{ego}} : \left| \mathbf{p}_{\text{ego}} - \mathbf{p} \right| \leq r_i \right\}, \; \mathbf{p} \in P_i, \; i = 1, 2,
\end{equation}
\begin{equation}
    P_1 = \{ \mathbf{p}_h \}, \; P_2 = \{ \mathbf{p}_{h, k} \}, \; 1 \leq h \leq H, \; 1 \leq k \leq K^\prime,
\end{equation}
\begin{equation}
    r_1 = r_{\text{ego}} + r_h + r_{\text{comfort}}, \quad r_2 = r_{\text{ego}} + r_h + \hat{\delta}_{h, k},
\end{equation}
where $D_1$ is the subarea around the current positions of humans and $D_2$ is the subarea around the predicted positions of humans. If the current center position of the ego robot $\mathbf{p}_{\text{ego}}$ is in either $D_1$ or $D_2$, an intrusion occurs. For the computation, we consider the distance between the center positions of agents and prediction points. $r_{\text{ego}}$ is the radius of the ego robot, $r_h$ is the radius of the $h$-th human, $r_{\text{comfort}}$ is the radius of the subarea around the current positions of humans, and $\hat{\delta}_{h, k}$ is the prediction uncertainty generated by DtACI for the $k$-th prediction point of the $h$-th human.

At each time step $t$, we iterate through all cost areas of all humans and calculate the maximum intrusion, denoted as $d_{\text{intru}, t}$. For an episode with a horizon of $T$, we have
\begin{equation} \label{eq:optimization_goal}
    \max_{\pi} \mathbb{E}_{\pi} \left[ \sum_{t=0}^{T} R_t(S_t, A_t) \right] \quad \text{s.t.} \quad \mathbb{E}_{\pi} \left[ \sum_{t=0}^{T} d_{\text{intru}, t} \right] = \tilde{d},
\end{equation}
where $\tilde{d}$ is a pre-defined threshold. 

We formulate the cost $C_t$ using the intrusions as $C_t(S_t, A_t) = \mu d_{\text{intru}, t}$,
where $\mu$ is a constant. Our reward includes three components: $R_{\text{success}}$, $R_{\text{collision}}$, and $R_{\text{potential}}$. $R_{\text{success}}$ is the reward for robot reaches the goal, $R_{\text{collision}}$ is for the robot collides with pedestrians, and $R_{\text{potential}}$ provides a dense reward that drives the ego robot to approach the goal, proportional to the distance the ego robot approaches the goal compared to the previous time step \cite{liu2023intention}.

In our work, we use the PPO Lagrangian \cite{ray2019benchmarking} for optimization. We set up two critics to compute the state value for reward and the state value for cost. The loss functions for the critics are defined as
\begin{align}
    l^R_t &= c_1 (V^R_{\theta_1}(S_t) - V_t^{\text{targ}, R})^2, \\
    l^C_t &= c_2 (V^C_{\theta_2}(S_t) - V_t^{\text{targ}, C})^2,
\end{align}
where $c_1$ and $c_2$ are constants, $V^R_{\theta_1}(S_t)$ and $V^C_{\theta_2}(S_t)$ are network-generated value estimates for reward and cost, respectively, and $V_t^{\text{targ}, R}$ and $V_t^{\text{targ}, C}$ are the corresponding target values.

As for the policy network, the action loss is similar to the form in PPO \cite{schulman2017proximal} except that we employ the combined advantage $\hat{A}^{\prime}_t = \frac{\hat{A}^R_t - \lambda \hat{A}^C_t}{1+\lambda}$.
The action loss function can then be written as
\begin{equation}
l_t^{\pi} = \hat{\mathbb{E}}_t \left[\min \left(r_t(\theta_3) \hat{A}^{\prime}_t, \operatorname{clip}\left(r_t(\theta_3), 1 - \epsilon, 1 + \epsilon\right) \hat{A}^{\prime}_t\right)\right],
\end{equation}
where $r_t(\theta_3)$ represents the change ratio between the updated and old policies, $\epsilon$ is the clip ratio, and $\hat{A}^{\prime}_t$ represents the estimated value advantage function at time step $t$.
Lastly, we update \(\lambda\) using gradient descent. The loss function \cite{ji2023omnisafe} for updating \(\lambda\) is defined as $l_t^{\lambda} = -\lambda(\bar{C} - \tilde{d}_C)$,
where \(\bar{C}\) is the mean episodic cost and \(\tilde{d}_C\) is the \textit{cost limit} proportional to $\tilde{d}$.

\section{Experiments}
\label{sec:experiments}
\subsection{Simulation Settings, Evaluation Metrics, and Implementation Details}
\label{sec:simulation_settings}
We adopt CrowdNav \cite{chen2019crowd}, one of the most widely adopted simulation environments for crowd navigation.
The environment consists of 20 humans and one robot in a \SI{12}{m} $\times$ \SI{12}{m} area, with randomized initial positions and goal locations.
The robot has a radius of \SI{0.2}{m}, while human radii are randomly sampled between \SI{0.3}{m} and \SI{0.5}{m}. The robot’s maximum speed is set to \SI{1.0}{m/s}. We adopt the standard evaluation metrics \cite{liu2023intention}, including success rate (SR), collision rate (CR), timeout rate (TR), navigation time (NT), path length (PL), intrusion time ratio (ITR), and social distance (SD). The detailed definitions can be found in Appendix \ref{sec:metrics}. For all of these metrics, SR reflects the effectiveness and CR evaluates the safety of algorithms. TR, NT, and PL measure the efficiency of the policy-generated paths, and ITR and SD measure the politeness of the paths. While different deployment contexts may prioritize different metrics, we emphasize SR and CR given their critical importance for safe robot deployment. We include our implementation details in Appendix \ref{sec:implementation_details}.

\begin{table*}[!tbp]
    \setlength{\tabcolsep}{1mm}
    \centering
    \caption{In-Distribution Test Results}
    \vspace{0.06cm}
    \fontsize{5.5}{6}\selectfont
    \resizebox{1.0\textwidth}{!}{
        \renewcommand{\arraystretch}{0.6}
       \begin{tabular}{
            m{1.3cm}<{\raggedright} | m{1.3cm}<{\centering} m{1.3cm}<{\centering} m{1.3cm}<{\centering} m{1.0cm}<{\centering} m{1.0cm}<{\centering} m{1.3cm}<{\centering} m{1.0cm}<{\centering} m{1.3cm}<{\centering}
        }
            \toprule
            Methods & \textbf{SR}$\uparrow$ & \textbf{CR}$\downarrow$ & \textbf{TR}$\downarrow$ & \textbf{NT}$\downarrow$ & \textbf{PL}$\downarrow$ & \textbf{ITR}$\downarrow$ & \textbf{SD}$\uparrow$ \\
            \midrule 
            SF & 15.60\% & 21.44\% & 62.96\% & 30.23 & 34.64 & 3.78\% & 0.42 \\
            ORCA & 67.84\% & 27.52\% & 4.64\% & 22.80 & 19.74 & \textbf{1.10}\% & \textbf{0.50} \\
            MPC  & 73.76\% & 25.52\% & 0.72\% & 19.09 & 20.88 & 11.91\% & 0.43\\
            SafeCrowdNav & 89.09$\pm$3.18\% & 10.91$\pm$3.18\% & \textbf{0.00$\pm$0.00}\% & 13.06$\pm$0.34 & \textbf{12.06$\pm$0.45} & 14.09$\pm$0.91\% & 0.39$\pm$0.01 \\
            CrowdNav++  & 86.11$\pm$0.61\% & 13.81$\pm$0.57\% & 0.00$\pm$0.14\% & 14.96$\pm$1.07 & 20.68$\pm$0.88 & 6.61$\pm$0.98\% & 0.43$\pm$0.01 \\
            \midrule                                       
            RL (w/o ACI) & 92.67$\pm$1.51\% & 7.33$\pm$1.51\% & \textbf{0.00$\pm$0.00}\% & \textbf{12.89$\pm$0.10} & 19.42$\pm$0.12 & 10.62$\pm$0.57\% & 0.39$\pm$0.01 \\
            RL (w/ ACI) & 94.08$\pm$1.18\% & 5.92$\pm$1.18\% & \textbf{0.00$\pm$0.00}\% & 13.35$\pm$0.26 & 19.88$\pm$0.25 & 8.81$\pm$0.67\% & 0.40$\pm$0.00 \\
            \midrule
            Ours (w/ CV) & 96.03$\pm$1.14\% & 3.73$\pm$1.24\% & 0.24$\pm$0.24\% & 17.88$\pm$0.60 & 24.51$\pm$0.76 & 2.40$\pm$0.22\% & 0.45$\pm$0.00 \\
            Ours (w/ GST) & \textbf{96.93$\pm$0.68}\% & \textbf{2.93$\pm$0.61}\% & 0.13$\pm$0.12\% & 17.54$\pm$0.86 & 24.27$\pm$0.85 & 2.72$\pm$0.16\% & 0.44$\pm$0.00 \\
            \bottomrule
        \end{tabular}
    }
    \label{tab:in_distribution_result}
\end{table*}

\begin{table*}[!tbp]
    \setlength{\tabcolsep}{1mm}
    \centering
    \caption{Out-of-Distribution Test Results}
    \vspace{0.06cm}
    \fontsize{5.5}{6}\selectfont
    \resizebox{1.0\textwidth}{!}{
        \renewcommand{\arraystretch}{0.6}
       \begin{tabular}{
            m{1.05cm}<{\centering} | m{1.2cm}<{\centering} | m{1.2cm}<{\centering} m{1.3cm}<{\centering} m{1.3cm}<{\centering} m{1.0cm}<{\centering} m{1.0cm}<{\centering} m{1.3cm}<{\centering} m{1.0cm}<{\centering} m{1.3cm}<{\centering}
        }
            \toprule
            Environments & Methods & \textbf{SR}$\uparrow$ & \textbf{CR}$\downarrow$ & \textbf{TR}$\downarrow$ & \textbf{NT}$\downarrow$ & \textbf{PL}$\downarrow$ & \textbf{ITR}$\downarrow$ & \textbf{SD}$\uparrow$ \\
            \midrule 
            \multirow{11}{*}{\shortstack{Rushing\\Humans}} & SF & 12.24\% & 19.12\% & 68.64\% & 32.06 & 36.15 & 5.31\% & 0.40 \\
            &ORCA & 60.32\% & 34.96\% & 4.72\% & 23.41 & 19.84 & \textbf{2.95} \% & \textbf{0.48} \\
            &MPC  & 49.76\% & 49.84\% & 0.40\% & 18.52 & 17.77 &19.70\% & 0.39 \\
            &SafeCrowdNav & 69.71$\pm$4.55\% & 30.29$\pm$4.55\% & \textbf{0.00$\pm$0.00}\% & 13.47$\pm$0.40 & \textbf{10.96$\pm$0.66} & 20.20$\pm$1.10\% & 0.37$\pm$0.00 \\
            &CrowdNav++ & 73.17$\pm$1.24\% & 26.67$\pm$1.40\% & 0.16$\pm$0.16\% & 15.43$\pm$1.63 & 19.89$\pm$1.12 & 12.38$\pm$1.48 & 0.39$\pm$0.00 \\
            &RL (w/o ACI) & 74.19$\pm$1.26\% & 25.81$\pm$1.26\% & \textbf{0.00$\pm$0.00}\% & \textbf{13.31$\pm$0.30} & 18.20$\pm$0.10 & 18.23$\pm$0.31\% & 0.37$\pm$0.00 \\
            &RL (w/ ACI) & 76.96$\pm$4.29\% & 23.04$\pm$4.29\% & \textbf{0.00$\pm$0.00}\% & 14.13$\pm$0.32 & 19.04$\pm$0.14 & 15.84$\pm$1.14\% & 0.37$\pm$0.01 \\
            &Ours (w/ CV) & 87.07$\pm$0.89\% & \textbf{12.75$\pm$1.21}\% & 0.19$\pm$0.32\% & 18.74$\pm$0.31 & 24.57$\pm$0.66 & 5.29$\pm$0.20\% & 0.40$\pm$0.00 \\
            &Ours (w/ GST) & \textbf{87.17$\pm$4.14}\% & \textbf{12.75$\pm$4.00}\% & 0.08$\pm$0.14\% & 18.32$\pm$1.01 & 24.04$\pm$0.85 & 6.82$\pm$1.36\% & 0.38$\pm$0.00 \\
            \midrule
            \multirow{11}{*}{\shortstack{SF Pedestrian\\Model}}&SF & 12.08\% & 6.72\% & 81.20\% & 29.76 & 40.56 & 1.60\% & 0.45 \\
            &ORCA & 92.56\% & 4.88\% & 2.56\% & 22.36 & 21.91 & \textbf{0.72}\% & \textbf{0.48} \\
            &MPC & 89.76\% & 10.00\% & 0.24\% & 17.07 & 20.58 & 8.58\% & 0.41 \\
            &SafeCrowdNav & 91.28$\pm$0.60\% & 8.72$\pm$0.60\% & \textbf{0.00$\pm$0.00}\% & \textbf{12.12$\pm$0.23} & \textbf{11.37$\pm$0.29} & 10.94$\pm$0.40\% & 0.39$\pm$0.00 \\
            &CrowdNav++ & 92.48$\pm$1.36\% & 7.52$\pm$1.36\% & \textbf{0.00$\pm$0.00}\% & 14.65$\pm$1.81 & 20.82$\pm$1.71 & 6.48$\pm$0.88\% & 0.41$\pm$0.00 \\
            &RL (w/o ACI) & 95.68$\pm$0.89\% & 4.32$\pm$0.89\% & \textbf{0.00$\pm$0.00}\% & 12.35$\pm$0.21 & 18.99$\pm$0.16 & 9.98$\pm$0.49\% & 0.39$\pm$0.00 \\
            &RL (w/ ACI) & 97.41$\pm$0.81\% & 2.59$\pm$0.81\% & \textbf{0.00$\pm$0.00}\% & 13.08$\pm$0.24 & 19.81$\pm$0.24 & 8.07$\pm$0.43\% & 0.40$\pm$0.01 \\
            &Ours (w/ CV) & 98.48$\pm$0.92\% & 1.39$\pm$0.82\% & 0.13$\pm$0.12\% & 19.02$\pm$0.45 & 25.84$\pm$0.54 & 2.04$\pm$0.25\% & 0.43$\pm$0.01 \\
            &Ours (w/ GST) & \textbf{98.96$\pm$0.52}\% & \textbf{1.04$\pm$0.52}\% & \textbf{0.00$\pm$0.00}\% & 18.18$\pm$0.84 & 25.05$\pm$0.79 & 2.66$\pm$0.45\% & 0.42$\pm$0.00 \\
            \midrule
            \multirow{11}{*}{\shortstack{Groups}}&SF & 2.56\% & 97.44\% & \textbf{0.00}\% & \textbf{9.41} & \textbf{9.91} & 23.59\% & 0.35 \\
            &ORCA & 49.44\% & 46.72\% & 3.84 & 22.98 & 19.37 & 20.46\% & 0.39 \\
            &MPC & 69.12\% & 29.76\% & 1.12\% & 20.48 & 22.78 & 24.08\% &0.37 \\
            &SafeCrowdNav & 75.73$\pm$2.01\% & 24.27$\pm$2.01\% & \textbf{0.00$\pm$0.00}\% & 16.29$\pm$0.44 & 13.44$\pm$0.64 & 26.78$\pm$1.42\% & 0.35$\pm$0.01 \\
            &CrowdNav++ & 81.84$\pm$15.17\% & 18.13$\pm$15.20\% & 0.03$\pm$0.05\% & 17.28$\pm$1.30 & 24.95$\pm$3.60 & 9.05$\pm$6.54\% & 0.39$\pm$0.01 \\
            &RL (w/o ACI) & 71.07$\pm$6.33\% & 28.93$\pm$6.33\% & \textbf{0.00$\pm$0.00}\% & 14.87$\pm$0.67 & 21.35$\pm$1.08 & 20.17$\pm$2.07\% & 0.36$\pm$0.01 \\
            &RL (w/ ACI) & 83.12$\pm$7.81\% & 16.88$\pm$7.81\% & \textbf{0.00$\pm$0.00}\% & 15.70$\pm$0.39 & 23.53$\pm$1.30 & 12.67$\pm$4.83\% & 0.37$\pm$0.01 \\
            &Ours (w/ CV)  & \textbf{94.51$\pm$4.20}\% & \textbf{5.41$\pm$4.23}\% & 0.08$\pm$0.14\% & 21.16$\pm$1.32 & 30.61$\pm$1.98 & \textbf{3.00$\pm$0.61}\% & \textbf{0.43$\pm$0.01} \\
            &Ours (w/ GST)  & 94.13$\pm$3.51\% & 5.71$\pm$3.31\% & 0.16$\pm$0.21\% & 20.19$\pm$0.69 & 29.69$\pm$0.81 & 3.55$\pm$1.21\% & 0.41$\pm$0.01 \\
            \bottomrule
        \end{tabular}
    }
    \label{tab:ood_result}
\end{table*}

\subsection{Baselines and Ablation Models}
Our baselines include optimal reciprocal collision avoidance (ORCA) \cite{van2008reciprocal}, social force (SF) \cite{helbing1995social}, model-predictive control (MPC), SafeCrowdNav \cite{xu2023safecrowdnav}, and CrowdNav++ \cite{liu2023intention}. ORCA and SF are classic rule-based algorithms in obstacle avoidance, MPC is an optimization-based algorithm, and SafeCrowdNav and CrowdNav++ represent the previous SOTA RL algorithms in crowd navigation. For MPC, we implement a cost function similar to that in \cite{dixit2023adaptive, huang2025interaction, lindemann2023safe}, which uses uncertainty as a reference in the cost formulation to improve safety. We test our method with the CV and the GST predictors, represented as Ours (w/ CV) and Ours (w/ GST), respectively. We also include RL (w/o ACI) and  RL (w/ ACI) as two ablation models. The two variants have the same network structures as Ours (w/ GST), but RL (w/o ACI) excludes uncertainty estimates from the input and does not employ CRL, while RL (w/ ACI) incorporates uncertainty estimates but does not apply CRL.

\begin{figure*}[!tbp]
    \centering
    \includegraphics[width=1.0\textwidth]{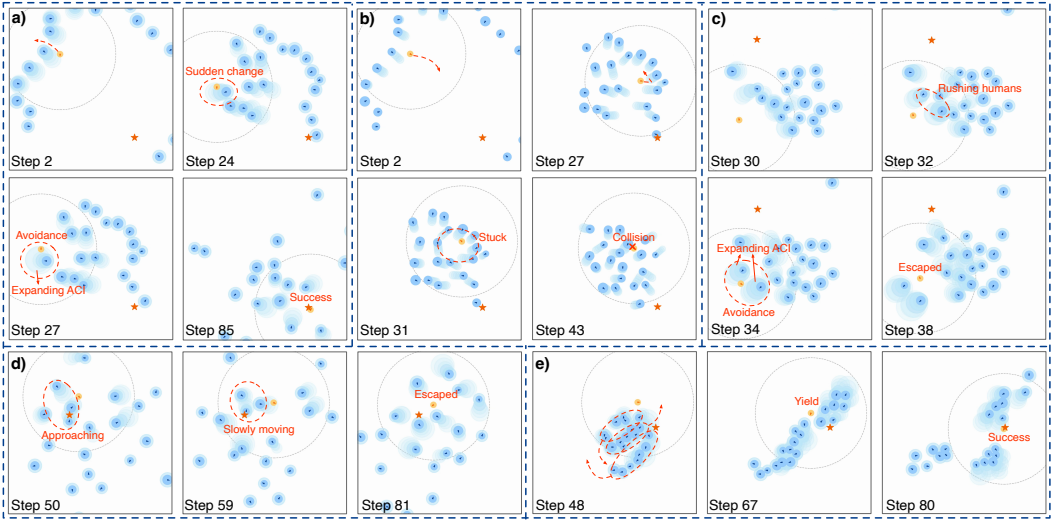}
    \caption{Test-case visualizations. Pedestrians are shown in blue, the robot in yellow, and the goal as an orange star. Light-blue circles show the quantified uncertainties around pedestrians. (a) Ours (w/ GST) successfully navigates to the goal in an in-distribution environment. (b) CrowdNav++ fails to complete the same episode; here, light-blue circles represent predicted trajectories rather than uncertainties. (c) Ours (w/ GST) in an OOD environment with rushing pedestrians. (d) Ours (w/ GST) in an OOD environment using the SF pedestrian model. (e) Ours (w/ GST) in an OOD environment with pedestrian groups.}
    \label{fig:QualitativeAnalysis}
\end{figure*}

\subsection{In-Distribution Test Results}
\label{sec:in-dist}
\textbf{Quantitative Analysis.} For all the test results in Table \ref{tab:in_distribution_result}-\ref{tab:ood_result}, we evaluate 1250 samples across 5 random test seeds and calculate the mean performance and standard deviations of models trained with 3 different training seeds. 
The test results under the same setting as the training environment are shown in Table \ref{tab:in_distribution_result}. From these results, we can see that SF, ORCA, and MPC have lower SR and higher CR compared to RL methods. When comparing all the RL-based methods, both Ours (w/ GST) and Ours (w/ CV) outperform other methods in safety metrics and ITR, indicating that our method can generate both safe and polite trajectories that cause minimal intrusions to pedestrians. Although our method generates paths with higher TR, NT, and PL compared to RL baselines and ablation models, this tradeoff is reasonable considering the overall improvement in effectiveness and safety. Notably, Ours (w/ GST) achieves over an 8.80\% higher SR, decreases collisions by more than 3.72 times, and reduces intrusions into the future pedestrian trajectories by more than 2.43 times compared to the two RL baselines.

\textbf{Qualitative Analysis.} We visualize the behaviors of Ours (w/ GST) and CrowdNav++ in the same episode in Fig. \ref{fig:QualitativeAnalysis}(a) and Fig. \ref{fig:QualitativeAnalysis}(b), respectively. At the beginning, CrowdNav++ decides to approach the goal directly. However, as the pedestrians move, they gradually surround the robot, which leads to an almost inevitable collision for CrowdNav++. 
In contrast, Ours (w/ GST) chooses to move the robot out of the crowds from the start. 
At step 24, Ours (w/ GST) rapidly reacts to a sudden change in human direction, and the adaptive expanding uncertainty area due to prediction errors helps the robot perform an avoidance maneuver. Additional qualitative examples are provided in our supplementary video.

\subsection{Out-of-Distribution Test Results}
\label{sec:ood}
\textbf{OOD Scenarios Mixed with Rushing Humans.} 
In this setting, we set 20\% of the human agents to have a maximum speed of \SI{2.0}{m/s}. From the results in Table \ref{tab:ood_result}, we observe that all methods exhibit degraded performance compared to in-distribution conditions.
Especially, SafeCrowdNav, CrowdNav++, RL (w/o ACI), and RL (w/ ACI) experience SR drops of 19.38\%, 12.94\%, 18.48\%, and 17.12\%, respectively. 
In contrast, Ours (w/ CV) and Ours (w/ GST) exhibit smaller SR drops of 8.96\% and 9.76\%, respectively. We visualize a scenario performed by Ours (w/ GST) in Fig. \ref{fig:QualitativeAnalysis}(c), where two rushing humans are moving toward the agent at step 32. Due to the GST predictor facing the challenges of OOD conditions, the uncertainty area generated by ACI becomes much larger. The robot then chooses to navigate between these two agents through the gap between the uncertainty areas at step 34 and successfully escapes to a safe area at step 38.

\textbf{OOD Scenarios with SF Pedestrian Model.}
In this setting, we change the behavior policy of all human agents from ORCA to SF. From the results in Table \ref{tab:ood_result}, all methods perform better than in in-distribution conditions.
Ours (w/ CV) and Ours (w/ GST) achieve almost perfect results in terms of SR, CR, and TR, implying that our methods \textit{adapt well} to OOD scenarios caused by different behavior models.
We visualize a scenario performed by Ours (w/ GST) in Fig. \ref{fig:QualitativeAnalysis}(d). When three humans approach each other, Ours (w/ GST) maintains a conservative distance from the crowds and successfully escapes afterward.

\textbf{OOD Scenarios with Group Dynamics.}
In this setting, pedestrians are clustered into cohesive groups that maintain tight intra-group spacing. As Table \ref{tab:ood_result} shows, every RL baseline and ablation variant suffers a noticeable performance drop. Especially, SafeCrowdNav and RL (w/ ACI) fall to SR below 76\%. The adaptability to grouped pedestrians of CrowdNav++ is highly sensitive to the training seed, with variance exceeding 15\%. By contrast, Ours (w/ CV) and Ours (w/ GST) retain SR above 94\% and CR below 6\%, while also displaying the smallest variance across seeds.  

\subsection{Other Test Results}
We include additional test results in Appendix \ref{sec:aci_effectiveness} to \ref{sec:visible_robot}, covering the effectiveness of ACI quantification, a convergence comparison between Ours (w/ GST) and RL (w/ ACI), the influence of different cost limits on policy aggressiveness, and cross-evaluations of algorithms trained in environments where humans either actively avoid or ignore the robot.

\subsection{Real-Robot Experiments}
We deploy our method on a ROSMASTER X3 robot with Mecanum wheels running ROS2, connected via router to a laptop equipped with a mobile NVIDIA RTX 3070 GPU. We apply the model to the robot without further fine-tuning, using only basic clipping and smoothing. Perception is performed with a 6 Hz RPLIDAR-A1 LiDAR, human detection via a pre-trained DR-SPAAM model \cite{jia2020dr}, tracking with SORT \cite{bewley2016simple}, and trajectory prediction with GST. We implement two movement modes: fixed-goal reaching and long-range navigation with a dynamically updated goal. The robot navigates outdoor environments and executes safe behaviors in both sparse and dense crowds. For more details, please refer to Appendix \ref{sec:ros2}-\ref{sec:compute_speed} and the supplementary video.

\section{CONCLUSION}
In this paper, we introduce an RL-based trajectory-planning framework that integrates conformal uncertainty into a CRL scheme to mitigate OOD performance degradation. Unlike conventional RL planners that overfit and falter under distribution shifts, our method dynamically leverages uncertainty estimates to adapt to velocity variations, policy changes, and transitions from individual to group dynamics. Extensive simulations demonstrate robust stability across diverse OOD scenarios, and real-world trials confirm the practical effectiveness of our approach.

\clearpage

\section{Limitations}
\label{sec:limitations} 
First, our work addresses the safety challenges of navigating dense, human-populated environments by combining uncertainty-aware prediction with constrained reinforcement learning. It can also handle static obstacles when they are modeled as collections of static agents, as in prior work \cite{chen2017decentralized}. However, to naturally handle obstacles of arbitrary shape, our approach would require additional perception and mapping capabilities, such as integrating SLAM and feeding multi-modal observations into the policy network. This represents a clear direction for future research.
The second limitation arises from perception errors (e.g., human miss-detections and the 2D LiDAR’s sensitivity to lighting). To mitigate these issues, we plan to augment our sensor suite with cameras, apply sensor-fusion techniques, and upgrade to a higher-frequency, more robust LiDAR, thereby adding safety redundancy.
Third, while our strategy effectively \textit{mitigates} OOD performance degradation, it does not yet \textit{eliminate} it, which is a great challenge in deep learning. Promising directions include adopting larger models (e.g., LLMs or VLMs) and advanced alignment methods, though these may introduce increased computational demands that could limit deployment on cost-effective robots.
Finally, although we have evaluated OOD performance in three representative scenarios: velocity variations, policy changes, and transitions from individual to group-level dynamics, we acknowledge that the space of possible distributional shifts is vast and cannot be exhaustively tested. We plan to conduct larger-scale real-world experiments once static-obstacle handling is resolved. These real-world experiments represent the most challenging OOD setting. We will also employ metrics such as the intervention rate to further compare our method against standard RL baselines.

\bibliography{example}

\clearpage
\appendix
\section*{Appendices for \emph{Towards Generalizable Safety in Crowd Navigation
via Conformal Uncertainty Handling}}

\section{Preliminaries}
\textbf{Adaptive Conformal Inference.}
Conformal methods can augment model predictions with a prediction set that is guaranteed to contain true values with a predefined coverage, enabling the quantification of uncertainties in a model-agnostic manner \cite{gibbs2024conformal}. Traditional split conformal prediction requires a calibration set and places high demands on the exchangeability between the test sample and the calibration samples. 
In contrast, adaptive conformal inference (ACI) can dynamically adjust its parameters to maintain coverage in an online and distribution-free manner \cite{gibbs2021adaptive}, making it appealing for time-sequential applications. Dynamically-tuned adaptive conformal inference (DtACI) \cite{gibbs2024conformal} further boosts the applicability and performance of ACI by running multiple prediction error estimators with different learning rates simultaneously. DtACI adaptively selects the best output based on its historical performance, eliminating the need to pre-acquire underlying data dynamics to achieve satisfying coverage.

\textbf{Constrained Reinforcement Learning.}
Constrained reinforcement learning (CRL) extends RL algorithms by incorporating constraints on the agent's behavior. Unlike traditional Markov decision process (MDP) settings, where agents learn behaviors only to maximize rewards, CRL is often formulated as a constrained Markov decision process (CMDP). At time step \( t \), an agent chooses an action \( A_t \) under state \( S_t \), receives a reward \( R_t \), and incurs a cost \( C_t \), after which the environment transitions to the next state \( S_{t+1} \). In a CMDP, the objective is not only to find an optimal policy that maximizes rewards but also to manage costs associated with certain actions or states, which may be defined as quantities related to safety in the context of social navigation. This is generally represented as \cite{ray2019benchmarking}:
\begin{equation}
    \pi^*=\arg \max _{\pi \in \Pi^C} J^R(\pi),
\end{equation}
where \( J^R(\pi) \) is a reward-based objective function, and \( \Pi^C \) is the feasible set of policies that satisfy the constraints added to the problem. The goal of CRL is to ensure that costs remain within pre-defined thresholds while maximizing reward.

\section{Conceptual Differences with CBF}
In CBFs, a safety function $h(x)$ and a safe set $\{x : h(x) > 0\}$ are used to encode safety. The condition $\dot{h}(x) \geq -\gamma h(x)$ is enforced during optimization to ensure the system remains in the safe set, constraining the \textit{derivative} of the safety function.
In contrast, our framework focuses on controlling the \textit{integral} of a risk-related function. Our cost function $C(S, A)$, analogous to $h(x)$, is accumulated over time as $\sum_{t=0}^{T} C_t(S_t, A_t)$, and we aim to constrain this cumulative cost below a small threshold $\tilde{d}_C$. 
Thus, while both approaches can shape the distribution of a safety-related function, the optimization mechanisms differ significantly.
Our methods integrate more closely with CRL and achieve expected safety.

\section{SF vs. ORCA}
\begin{figure}[!tbp]
	\centering
	\includegraphics[width=1.00\columnwidth]{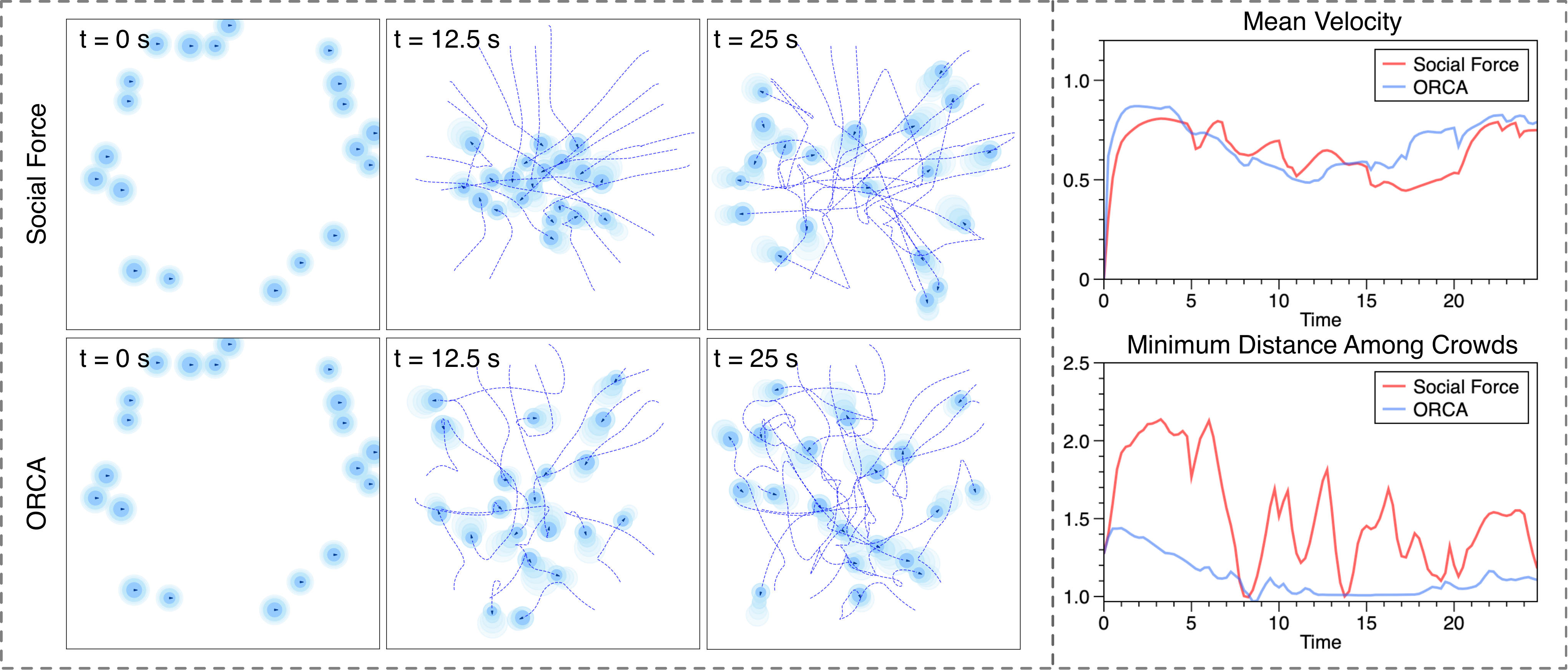}
	\caption{Comparison of SF and ORCA.
    }
	\label{fig:sforca}
\end{figure}

We plotted human trajectories generated using ORCA and SF dynamics in Fig. \ref{fig:sforca}, along with plots of minimal distances between individuals and mean velocities. Note that in our simulation environments, humans may sample a new goal every five steps with a probability of 50\%, which changes their moving directions and velocities. From the results, ORCA pedestrians exhibit sharper turns and smaller minimal distances between individuals.

\section{Evaluation Metrics}
\label{sec:metrics}
The evaluation metrics used in our paper are introduced in detail as follows:
\begin{itemize}
    \item \textit{Success Rate (SR):} 
    SR measures the ratio of the number of successful episodes (i.e., the robot reaching the goal within the time limit of \SI{50}{s})
    to the total number of test episodes, i.e., 
    \( \text{SR} = N_{\text{success}} / N_{\text{total}} \).

    \item \textit{Collision Rate (CR:}
    CR measures the ratio of episodes with at least one collision
    to the total number of test episodes, i.e.,
    \( \text{CR} = N_{\text{collision}} / N_{\text{total}} \).

    \item \textit{Timeout Rate (TR):}
    TR measures the ratio of episodes that fail to reach the goal before the time limit
    to the total number of test episodes, i.e.,
    \( \text{TR} = N_{\text{timeout}} / N_{\text{total}} \).

    \item \textit{Navigation Time (NT):}
    NT is the average time required for the robot to reach the goal (computed only over \textit{successful} episodes), i.e.,
    \( \text{NT} = \bigl(1 / N_{\text{success}}\bigr) \sum_{k=1}^{N_{\text{success}}} T_k \),
    where \(T_k\) is the time the robot took to reach the goal in the \(k\)-th successful episode.

    \item \textit{Path Length (PL):}
    PL is the total distance traveled by the robot in an episode, accumulated step by step.
    We report the average path length across all episodes (including collisions and timeouts), i.e.,
    \( \text{PL} = \bigl(1 / N_{\text{total}}\bigr) \sum_{i=1}^{N_{\text{total}}} \text{dist}_i \),
    where \(\text{dist}_i\) is the total distance traveled in episode \(i\).

    \item \textit{Intrusion Time Ratio (ITR):}
    ITR is the fraction of time steps in an episode
    where the robot intrudes into the ground-truth future positions of humans (\textit{Danger} event triggered).
    Its final value is an average over all episodes:
    \begin{equation}
        \text{ITR} 
        = \frac{1}{N_{\text{total}}} \sum_{i=1}^{N_{\text{total}}}
        \Bigl(\frac{\#\,\text{danger steps in episode } i}{\text{total steps in episode } i} 
        \times 100\%\Bigr).
    \end{equation}

    \item \textit{Social Distance (SD):} SD is the average of the minimal distances between the robot and any human,
    computed only at the ``danger steps.'' We log the minimal distance whenever a \textit{Danger} event occurs, and then take the average over all episodes.
\end{itemize}

\begin{figure}[!tbp]
	\centering
	\includegraphics[width=0.5\columnwidth]{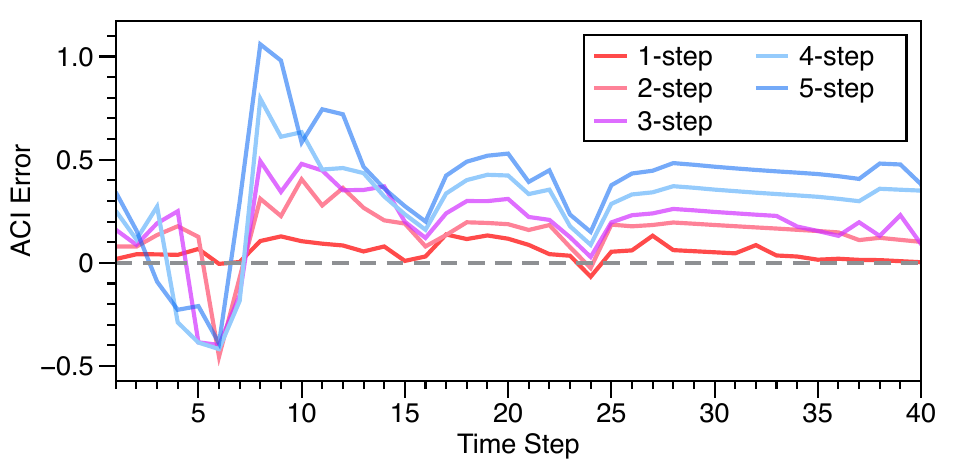}
	\caption{Visualization of ACI errors (i.e., estimated prediction error minus actual prediction error) for one pedestrian's five prediction steps during the time period it is within the observable area of the robot. ACI provides valid coverage when the ACI error is greater than 0.
    }
	\label{fig:ACIError}
\end{figure}

\section{Implementation Details}
\label{sec:implementation_details}
Under each ablation setting, we train our models using three different random seeds on an NVIDIA RTX 4090 GPU, keeping all other hyperparameters consistent except those we intend to compare. Key common settings include: 
\begin{itemize}
    \item All models use human predictions as part of the input observations and employ a pre-trained GST predictor with fixed parameters to generate five steps of human trajectory predictions, except for Ours (w/ CV).
    \item For the training parameters, we leverage the parallel training feature of PPO by and running 128 parallel environments. The learning rate is set to $3\times10^{-5}$, and the clip parameter is set to 0.08 for both the actor and the reward critic. The learning rate for the cost critic is set to $1.5\times10^{-5}$. We run $2\times10^7$ training steps to obtain the final policy model for testing and real-robot deployment. In our training, varying entropy is removed, and the action noise, which is normally distributed, is only added during training and removed during testing and deployment.
    \item We set the radius of the cost subareas around current positions of humans, \(r_{\text{comfort}}\), to \SI{0.25}{m} and calculate costs based on the intrusions into the subarea and the first two prediction uncertainty areas, while inputting predictions for all five future steps into the policy network with a coverage parameter \(\alpha = 0.1\), corresponding to 90\% coverage in prediction errors.
    \item For DtACI hyperparameters, the initial prediction errors are set to \SI{0.1}{m}, \SI{0.2}{m}, \SI{0.3}{m}, \SI{0.4}{m}, and \SI{0.5}{m} for 1- to 5-step-ahead predictions, respectively. We employ three error estimators with learning rates \(\gamma\) of 0.05, 0.1, and 0.2 for each DtACI estimator.
    \item For the specific reward and cost functions, we set the success reward $R_{\text{success}}$ to $+10$, the collision penalty $R_{\text{collision}}$ to $-20$, and the dense potential reward that guides the robot's movement to $2 \Delta d_{\text{forward}}$, where \(\Delta d_{\text{forward}}\) measures the distance the robot moves toward the goal compared to the previous step. The cost \(C_t(S_t, A_t)\) is defined as $2.5  d_{\text{intru}, t}$, where \(d_{\text{intru}, t}\) represents the maximum intrusion into the cost areas of the current human positions and the first two prediction uncertainty areas.
\end{itemize}

\section{Tradeoffs of Different Methods}
From the results in Tables \ref{tab:in_distribution_result}-\ref{tab:ood_result}, we observe that: (1) SF, ORCA, and MPC show poor performance across multiple metrics, particularly in collision avoidance and goal-reaching capabilities. (2) CrowdNav++, RL (w/o ACI), and RL (w/ ACI) achieve reasonable goal-reaching rates and safety performance. (3) SafeCrowdNav achieves comparable task completion and safety performance with shorter path lengths, but exhibits higher intrusion rates, indicating less socially-aware behavior. (4) Our method achieves the highest goal-reaching rates with the lowest collision rates and fewest intrusions, at the cost of longer paths and navigation time, since our framework prioritizes safety over efficiency. However, this safety-efficiency tradeoff can be easily adjusted via the cost limit parameter for different deployment scenarios.

\section{ACI Effectiveness}
\label{sec:aci_effectiveness}
To validate the effectiveness of ACI in covering actual prediction errors, we visualize the ACI errors of one human in Fig. \ref{fig:ACIError}. At the beginning of this trajectory, the ACI errors for multi-step predictions are large because the GST predictor lacks sufficient information to accurately predict human positions, leading to large actual prediction errors that our initial ACI values do not cover properly. However, after several steps, ACI quickly adapts to the actual prediction error and achieves adequate coverage. Additionally, if the coverage remains sufficient, the ACI error will decrease to ensure that the uncertainty estimation is not too conservative. We also extract human trajectories and the corresponding ACI predictions to evaluate the quantitative coverage performance. The coverage rates for 1-, 2-, 3-, 4-, and 5-step predictions are 90.21\%, 91.53\%, 92.34\%, 90.78\%, and 90.55\%, respectively, using the parameters specified in our approach with $\alpha = 0.1$, which theoretically corresponds to 90\% coverage.

\section{Convergence Analysis}
\begin{figure}[!tbp]
	\centering
	\includegraphics[width=0.8\columnwidth]{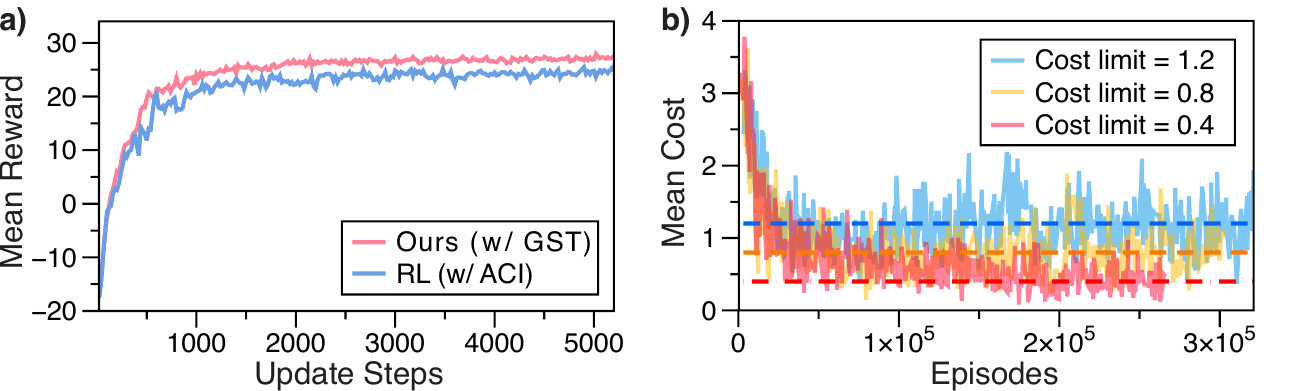}
	\caption{Convergence analysis of our method. (a) The learning curves of Ours (w/ GST) and RL (w/ACI). Ours (w/ GST) shows faster convergence with higher rewards. (b) The cost curves of Ours (w/ GST) with different cost limits. The average costs across episodes can approximately approach the predefined cost limits, which are shown by the dashed lines.
    }
	\label{fig:curves}
\end{figure}

We present the learning curves of Ours (w/ GST) in Fig. \ref{fig:curves}(a), alongside RL (w/ ACI) for comparison. The results indicate that Ours (w/ GST) demonstrates a smoother learning process and achieves higher rewards. Furthermore, as shown in Fig. \ref{fig:curves}(b), the average episodic costs of Ours (w/ GST) converge to different cost limit values. This provides an effective mechanism for adjusting the aggressiveness of robot policies, as further discussed in Appendix \ref{sec:aggressiveness}.

\section{Tuning Aggressiveness of Trajectories with Cost Limits}
\label{sec:aggressiveness}
Although the main results presented in our paper focus on low cost limits to ensure safe navigation through crowds, we also observe interesting testing results with varying cost limits that allow for tuning the aggressiveness of trajectories, as shown in Table \ref{tab:tune_aggressiveness}. These results are obtained by testing across five seeds, with 250 test samples for each seed, using models trained with the same training seed. We find that as the cost limits increase, the trajectories become gradually more aggressive, reflected by an increase in ITR. This increased aggressiveness also improves efficiency, as indicated by reductions in NT and PL. However, it comes at the expense of key metrics such as SR and CR. Therefore, for a single best policy, we recommend keeping the cost limit as low as possible, provided the policy can still converge.

\begin{table}[!tbp]
    \setlength{\tabcolsep}{0.4mm}
    \centering
    \caption{Performance of Ours (w/ GST) Under Different Cost Limits}
    \fontsize{5.5}{6}\selectfont
    \resizebox{0.6\textwidth}{!}{
        \renewcommand{\arraystretch}{0.6}
       \begin{tabular}{
            m{0.8cm}<{\centering} | m{0.8cm}<{\centering} m{0.6cm}<{\centering} m{0.6cm}<{\centering} m{0.6cm}<{\centering} m{0.6cm}<{\centering} m{0.6cm}<{\centering} m{0.6cm}<{\centering} m{0.6cm}<{\centering}
        }
            \toprule
            Cost limit & \textbf{SR}$\uparrow$ & \textbf{CR}$\downarrow$ & \textbf{TR}$\downarrow$ & \textbf{NT}$\downarrow$ & \textbf{PL}$\downarrow$ & \textbf{ITR}$\downarrow$ & \textbf{SD}$\uparrow$ \\
            \midrule 
            0.4 & 96.96\% & 2.80\% & 0.24\% & 18.52 & 25.24 & 2.73\% & 0.43 \\
            0.6 & 96.72\% & 3.04\% & 0.24\% & 16.16 & 22.96 & 3.59\% & 0.44 \\
            0.8 & 94.56\% & 5.44\% & 0.00\% & 13.60 & 20.33 & 5.61\% & 0.43 \\
            1.0 & 94.32\% & 5.68\% & 0.00\% & 13.53 & 20.19 & 6.62\% & 0.42 \\
            1.2 & 93.76\% & 6.24\% & 0.00\% & 12.97 & 19.45 & 7.31\% & 0.41 \\
            \bottomrule
        \end{tabular}
    }
    \label{tab:tune_aggressiveness}
\end{table}

\section{Visible Robot Settings}
\label{sec:visible_robot}
For the main results presented in our paper, we train the RL baselines, ablation models, and our methods in the CrowdNav environments where humans will not actively avoid the robot, which is a common practice in prior work \cite{liu2023intention, xu2023safecrowdnav}. To validate the effectiveness of this setting, we conduct a cross-evaluation on models trained in environments where humans either actively avoid or ignore the robot, as summarized in Tables~\ref{tab:visible_CR}-\ref{tab:visible_SR}. Specifically, we evaluate the models under five different testing seeds, each with 250 test samples, for three models trained under three distinct training seeds. From these results, we observe that although models trained in the visible robot setting achieve low collision rates in their in-distribution settings, they fail to generalize when humans do not react to the robot. In such scenarios, nearly half of the test cases result in collisions. In contrast, models trained in the invisible robot setting maintain low collision rates in both testing scenarios and, surprisingly, exhibit even lower collision rates when the robot is visible than those trained in the visible robot setting. In real-world situations, some pedestrians may ignore the robot (e.g., when they are using their mobile phones while walking). This mixture of reactive and non-reactive pedestrians poses significant challenges for models that assume all humans will actively avoid the robot. Thus, training in an environment where humans do not react to the robot proves to be more robust, as it better accounts for scenarios in which pedestrians may fail to notice the robot.

\begin{table}[!tbp]
    \centering
    \caption{Cross-Validation of Ours (w/ GST) for Robot Visibility Settings Using Success Rates}
    \fontsize{5.5}{6}\selectfont
    \resizebox{0.40\textwidth}{!}{
    \renewcommand{\arraystretch}{1.0}
    \begin{tabular}{%
        m{1.2cm}<{\centering} | 
        m{1.2cm}<{\centering} 
        m{1.2cm}<{\centering}
    }
    \toprule
    \diagbox[width=1.2cm]{\textbf{Test}}{\textbf{Train}} 
        & \textbf{Invisible} 
        & \textbf{Visible} \\
    \midrule
    \textbf{Invisible} & \textbf{96.93$\pm$0.68}\% & 50.48$\pm$0.49\%\\
    \midrule
    \textbf{Visible}   &  \textbf{99.92$\pm$0.14}\%& 99.07$\pm$0.76\%\\
    \bottomrule
    \end{tabular}
    } 
    \label{tab:visible_CR}
\end{table}

\begin{table}[!tbp]
    \centering
    \caption{Cross-Validation of Ours (w/ GST) for Robot Visibility Settings Using Collision Rate}
    \fontsize{5.5}{6}\selectfont
    \resizebox{0.40\textwidth}{!}{
    \renewcommand{\arraystretch}{1.0} 
    \begin{tabular}{
        m{1.2cm}<{\centering} | 
        m{1.2cm}<{\centering} 
        m{1.2cm}<{\centering} 
    }
    \toprule
    \diagbox[width=1.2cm]{\textbf{Test}}{\textbf{Train}} 
        & \textbf{Invisible} 
        & \textbf{Visible} \\
    \midrule
    \textbf{Invisible} & \textbf{2.93$\pm$0.61}\% & 49.52$\pm$0.49\%\\
    \midrule
    \textbf{Visible}   &  \textbf{0.08$\pm$0.14}\%& 0.93$\pm$0.76\%\\
    \bottomrule
    \end{tabular}
    }
    \label{tab:visible_SR}
\end{table}

\section{ROS2 System for Real Robot Deployment}
\label{sec:ros2}
We develop a full ROS2 system from perception to decision making, including four main nodes:
\begin{itemize}
    \item \textit{Detector:} This node employs a pretrained DR-SPAAM model \cite{jia2020dr} to detect human agents from 2D LiDAR point clouds. The model leverages a lightweight neural network, enabling real-time detection on resource-constrained devices. According to the original paper, it achieves an accuracy metric of $AP_{0.5} = 70.3\%$.
    \item \textit{Tracker:} This node uses the SORT \cite{bewley2016simple} algorithm for tracking, which simply aims at assigning indices to detected agents. To generate consistent tracking results even when receiving noisy detections, we assign large values to the measurement noise covariance matrix \(\mathbf{R}\) and the process noise covariance matrix \(\mathbf{Q}\) of the underlying Kalman filter to accommodate detection jitter. We also set a higher noise coefficient for the position and scale components of the Kalman filter and a smaller value for the velocity components to allow more flexibility in position and scale updates. In the initial state covariance matrix \(\mathbf{P}\), the velocity components are given a large uncertainty, and the entire matrix is scaled accordingly so the filter can quickly adapt during early tracking. Besides, in the SORT tracker, we only keep trajectories whose ratio of valid detections over the entire detected period is above 0.9, filtering out spurious or short-lived detections. At each timestep, we run the Kalman filter's inference step to infer the positions of tracked agents even when they are not correctly detected in certain frames.
    \item \textit{Predictor:} This node uses the GST \cite{huang2021learning} model as the trajectory planner, integrating the DtACI module to directly apply uncertainty quantification after obtaining prediction lines for the pedestrians. The parameters and model weights are consistent with those used during training.
    \item \textit{Decider:} This node receives information from the upper-level nodes and determines the output action commands ($v_x$ and $v_y$ for velocities) for the controller. We utilize the controller integrated into the ROSMASTER X3 system by publishing \verb|\cmd_vel|. The decider supports three modes: goal-reaching mode, long-range navigation mode, and manual control mode. The selection of these modes is managed by another utility node, \textit{Command Listener}, which listens to user input and communicates the commands to the decider.
\end{itemize}

\section{Real Robot Experiments} \label{sec:real_experiments}
We deploy our approach on a ROSMASTER X3 robot equipped with Mecanum wheels, enabling flexible movement and independent control of $v_x$ and $v_y$. 
The robot connects to a laptop with an Nvidia RTX 3070 GPU (mobile version) via a router. 
Our model is trained in CrowdNav simulations with holonomic dynamics and is deployed directly to the robot system \textit{without} further fine-tuning. 

The robot uses a 2D LiDAR (RPLIDAR-A1) for human detection via a learning-based detector. The LiDAR operates at a scanning frequency of approximately \SI{6}{Hz}, which also limits the frequencies of the tracking, prediction, and decision making modules. Despite this, our experiments demonstrate that the robot makes effective navigation decisions in crowded environments, indicating that the system generalizes well to cost-efficient robots. Human detection is done with a pre-trained DR-SPAAM model \cite{jia2020dr}, tracking is done with SORT \cite{bewley2016simple}, prediction is done with the GST model \cite{huang2021learning}, and decision making is powered by Ours (w/GST). For further details about our system, please refer to Appendix \ref{sec:ros2}, and computational speed analyses for the core functions of each module are provided in Appendix \ref{sec:compute_speed}.

We conduct the experiments on a large outdoor terrace covering an area greater than \SI{15}{m} $\times$ \SI{20}{m}. The robot operates in two movement modes:
\begin{itemize}
    \item \textit{Goal-reaching mode:} The robot navigates to a predetermined target point, which is the original setting in the standard CrowdNav training phase.
    \item \textit{Long-range navigation mode:} The goal of the robot is dynamically updated based on its movement. This approach addresses the limitation that, during training, the goal inputs to the policy networks are constrained to a limited range. As a result, distant goals beyond \SI{12}{m} may negatively impact policy performance. By introducing a continuously updated moving target, we show that the robot can reliably navigate distances exceeding \SI{20}{m} through experiments. This method has the potential to be extended to even longer distances if the system is fully integrated onboard in the future.
\end{itemize}

\begin{figure*}[!tbp]
	\centering
	\includegraphics[width=1.0\textwidth]{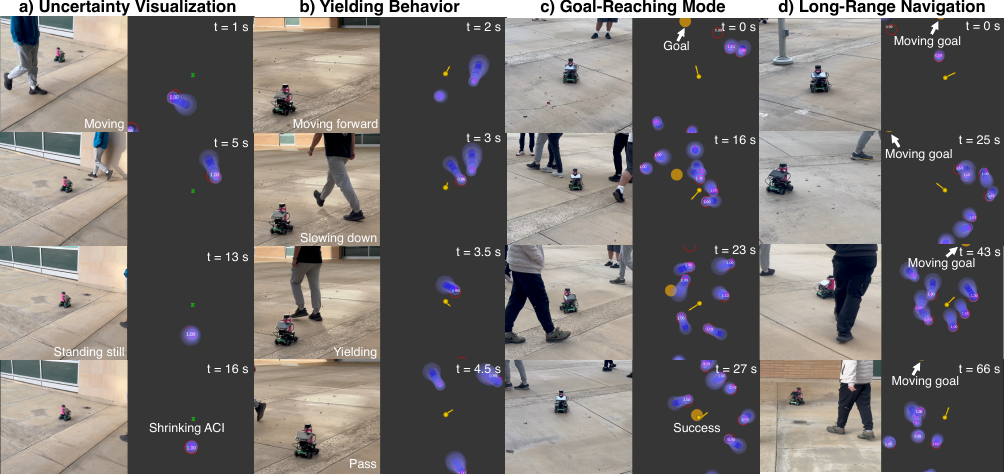}
	\caption{We deploy our methods on a ROSMASTER X3 with Mecanum wheels using the ROS2 system. For the four subplots, the left sides display photos taken from the experiments, and the right sides show visualizations in RViz. In the RViz visualizations, the red circles represent detection results, and the white numbers inside the red circles indicate the output probabilities of the detection model. The purple numbers correspond to the indices generated by the tracker. The prediction lines are shown in blue, and the prediction uncertainties are depicted in semi-transparent light blue. In (a), the green robot represents the robot’s current location. In (b)-(d), where the decision node is enabled, the yellow sphere indicates the robot’s position, and the yellow arrow represents the command output from the decider node. The orange circle indicates the goal position. (a) In the uncertainty visualization, the human initially walks around the robot and then stands still behind it. The uncertainty area adjusts dynamically based on the prediction accuracy. (b) The robot equipped with Ours (w/ GST) demonstrates stable yielding behavior when interacting with humans. (c) In goal-reaching mode, the robot navigates through crowds and successfully reaches its goal. (d) In long-range navigation mode, the moving goal consistently guides the robot’s movement.
    }
	\label{fig:experiments}
\end{figure*}

We show several representative real-world testing scenarios and results in Fig. \ref{fig:experiments}.
Specifically, we first visualize the prediction uncertainties using RViz in Fig. \ref{fig:experiments}(a). Before $t= \SI{13}{s}$, the human walks around the robot, and the robot successfully detects the human's position while visualizing the predicted trajectory and the associated uncertainty area. 
Since the prediction model does not fully capture the human's circular walking pattern, the long-horizon prediction uncertainty area is significantly larger than that of short-horizon predictions. After $t= \SI{13}{s}$, the human stands still behind the robot. Initially, the prediction uncertainty area is large because the uncertainty results are carried over from iterative processes. Gradually, the uncertainty area shrinks to reflect more precise prediction uncertainties. After approximately 3 seconds, the uncertainty area stabilizes and reduces to almost zero. This process demonstrates that by using DtACI, the system can effectively adapt to different conditions and dynamically reflect prediction uncertainties. This enables safer decision making without being overly conservative.

In Fig. \ref{fig:experiments}(b), by allowing the robot to interact with a single agent, we observe that the robot can generate stable yielding behaviors. At $t= \SI{2}{s}$, the robot detects that a human is coming along a path that will intersect with its original moving trajectory. It begins by slowing down to reduce the risk of a collision ($t= \SI{3}{s}$). At $t= \SI{3.5}{s}$, the robot adjusts its direction, moving towards the backside area of the human to avoid intrusions. At $t= \SI{4.5}{s}$, the robot successfully passes.

In Fig. \ref{fig:experiments}(c), the robot navigates in goal-reaching mode while interacting with humans along its path. We observe that the robot encounters very dense crowds at $t = \SI{16}{s}$ and $t = \SI{23}{s}$. Throughout this process, the robot makes appropriate decisions to minimize intrusions into humans’ trajectories while progressing toward its goal, and it ultimately reaches the goal despite the dense human interactions.

In Fig. \ref{fig:experiments}(d), we implement the long-range navigation mode, enabling the robot to navigate distances beyond those encountered during training. In this mode, the moving goal is dynamically updated to remain \SI{5}{m} ahead of the robot’s position along the longitudinal axis, while its lateral position remains fixed. Under these conditions, the robot moves forward while actively avoiding dynamic obstacles, as demonstrated at $t = \SI{25}{s}$. Although the robot temporarily deviates laterally to avoid collisions, the fixed lateral component of the moving goal serves as a stable reference, helping the robot stay within a reasonable lateral range.

Overall, our system demonstrates robust decision making capabilities in crowded environments. By quantifying prediction uncertainty and incorporating it into behavior guidance, the robot develops more adaptive spatial awareness and exhibits consistent, safe, and socially aware behaviors.

\section{Computational Speed Analysis}
\label{sec:compute_speed}
\begin{figure}[!tbp]
	\centering
	\includegraphics[width=0.5\columnwidth]{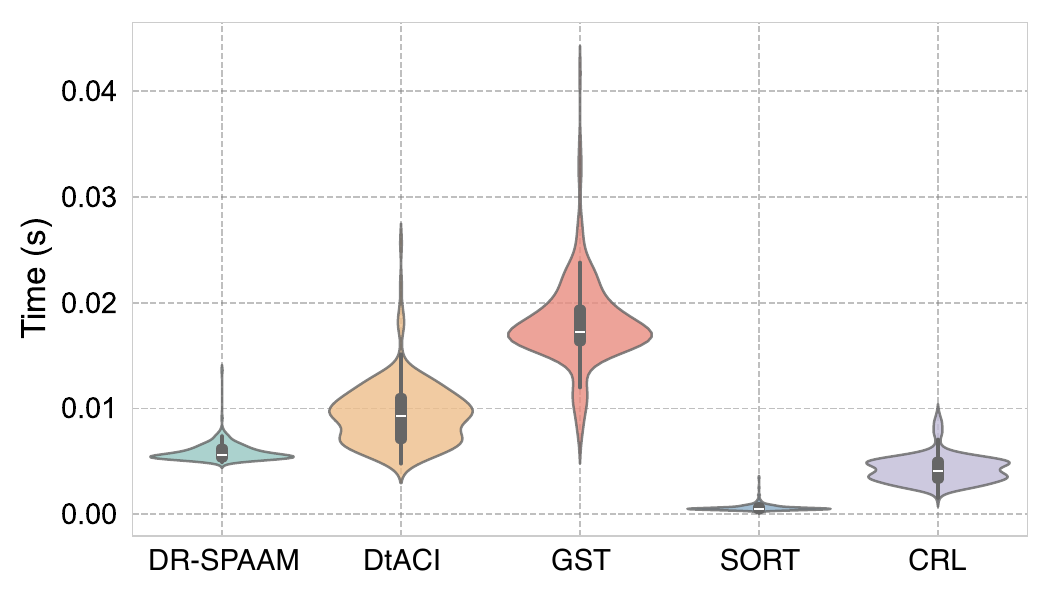}
	\caption{Visualizations of computational time distributions for five key modules of our ROS2 system. The data samples were collected during experiments in dense human environments. The shape of each violin plot represents the density of computation times, with wider areas indicating a higher concentration of values. The white dots at the center of each shape denote the mean computation time for the respective module.
    }
	\label{fig:computational_time}
\end{figure}

We log the time consumption of key functions across different nodes, including the inference times of the detection model (DR-SPAAM), prediction model (GST), and the policy network (CRL), as well as the update times for the tracker (SORT) and uncertainty quantifier (DtACI) in dense crowd experiments, and visualize the results in Fig. \ref{fig:computational_time}. The visualization shows that GST has the longest average computation time and the largest variance. In contrast, DR-SPAAM and SORT maintain low and stable computation times, making them efficient for real-time applications. While DtACI has relatively high computational costs due to its pure Python implementation, its average time remains below \SI{0.01}{s}, well within acceptable limits. This overhead can be greatly reduced by transitioning to more efficient languages like C++. Despite this, the worst-case computation time for all main modules stays below 0.1 seconds, meeting and exceeding our system’s decision frequency requirements.

\end{document}